\documentclass{bmvc2k}
\usepackage{algorithm}
\usepackage{algorithmic}

\title{Training DNNs in O(1) memory with MEM-DFA using Random Matrices}

\addauthor{Tien Chu}{chutien@protonmail.com}{1}
\addauthor{Kamil Mykitiuk}{k.mykitiuk@student.uw.edu.pl}{1}
\addauthor{Miron Szewczyk}{mj.szewczyk2@student.uw.edu.pl}{1}
\addauthor{Adam Wiktor}{adam.wiktor97@gmail.com}{1}
\addauthor{Zbigniew Wojna}{zbigniewwojna@gmail.com}{2}
\addinstitution{Warsaw University}
\addinstitution{Tensorflight, Inc.}

\runninghead{Chu et al.}{Training DNNs in O(1) memory with MEM-DFA}

\newcommand{\pder}[2]{\frac{\delta #1}{\delta #2}}
\begin{document}

\maketitle

\begin{abstract}
This work presents a method for reducing memory consumption to a constant complexity when training deep neural networks. The algorithm is based on the more biologically plausible alternatives of the backpropagation (BP): direct feedback alignment (DFA) and feedback alignment (FA), which use random matrices to propagate error. The proposed method, memory-efficient direct feedback alignment (MEM-DFA), uses higher independence of layers in DFA and allows avoiding storing at once all activation vectors, unlike standard BP, FA, and DFA. Thus, our algorithm's memory usage is constant regardless of the number of layers in a neural network. The method increases the computational cost only by a constant factor of one extra forward pass.

The MEM-DFA, BP, FA, and DFA were evaluated along with their memory profiles on MNIST and CIFAR-10 datasets on various neural network models. Our experiments agree with our theoretical results and show a significant decrease in the memory cost of MEM-DFA compared to the other algorithms.\end{abstract}

\section{Introduction}
State of the art CNN methods strive for high memory usage as in many scenarios, the deeper networks with lager inputs achieve better results. This is exemplified by many successful models such as ResNet \cite{He2016DeepRL}, Inception \cite{Szegedy2016RethinkingTI}, or SENet \cite{Hu2018SqueezeandExcitationN}.

The backpropagation seems to be an irreplaceable algorithm when it comes to computational cost. However, it imposes substantial memory costs. Apart from storing the model's parameters, the backpropagation stacks all intermediate activation vectors in the forward pass, necessary for gradient calculations during the backward pass. 

The memory problem becomes even more apparent for tasks involving processing high-resolution images. For example, the hardware limitations require a decrease in the quality of images, train the model in small batches, reduce the model's size, or increase the training time. 

It is believed the backpropagation is biologically implausible to perform in the brain \cite{random}. Feedback alignment \cite{random, liao2016important} is an alternative method that uses random matrices to propagate error instead of transposed weight matrices. Direct feedback alignment \cite{random2, random3}, on the other hand, propagates error from the last layer directly to each layer through the random matrix. 

This work proposes a method for reducing memory consumption called MEM-DFA. It is preceded by a detailed description of the backpropagation, FA and DFA. Here, we present the results from six experiments with different models on MNIST and CIFAR-10 datasets. We discuss the results and derive conclusions.

The contributions of this work are the following:
\begin{itemize}
    \item Review and analyze memory usage footprint of BP, FA, DFA, and MEM-DFA algorithms.
    \item Propose a new method for training DNNs called MEM-DFA that consumes O(1) memory regardless of the number of layers.
\end{itemize}

\section{Related works}
\subsection{Memory optimization}
The first group of related works addresses the memory optimization of neural networks. We can divide them into three categories:
\begin{itemize}
\item reducing the number of stored activation vectors at the peak by applying checkpointing method \cite{throughtime, sublinear, reforwarding, checkpoinint_openai},
\item neural networks' architecture modifications to decrease the number of activation vectors required in the backpropagation \cite{revnet, memcnn, batchnorm},
\item reducing the number of bits used to represent activation vectors \cite{lowerprecision, lowerprecision2, binarized},
\end{itemize}

The checkpointing method stores a subset of the activations and recalculates the missing activations during backpropagation. This is a trade-off between memory and computational cost. Depending on the algorithm of selection of checkpoints, the memory usage ranges from linear to constant, but the time complexity ranges from linear to quadratic.

Merging batch normalization with activation function \cite{batchnorm} is an example of reducing the number of sored activation vectors to decrease memory usage. Another example is a modification of ResNets, called RevNets, by introducing reversible blocks in place of residual blocks \cite{revnet, memcnn}.

Reducing the number of bits representing values of matrices in a neural network can decrease memory usage and increase calculations speed \cite{lowerprecision, lowerprecision2}. In \cite{binarized}, the authors trained a neural network using only one-bit values.

\subsection{Biologically motivated methods with random matrices}
This paper is inspired by the line of work trying to overcome backpropagation's biological implausibility. Feedback alignment \cite{random, liao2016important} and direct feedback alignment \cite{random2} relax information constraints of backpropagation by using random matrices to propagate error. Both FA and DFA showed promising results on the MNIST dataset achieving close to backpropagation accuracy \cite{random, liao2016important, random2}. FA algorithm in \cite{liao2016important} achieves even better results by preserving the signs of the matrices. In \cite{gilmer2017explaining}, the authors try to explain the effectiveness and limitations of DFA.

The works \cite{xiao2018biologically, bartunov2018assessing, moskovitz2018feedback, crafton2019direct} try to scale FA, DFA, and other biologically plausible algorithms such as Target Propagation \cite{lee2015difference} on large datasets. The authors of \cite{xiao2018biologically} show that preserving sign symmetry as in \cite{liao2016important} allows scaling FA to satisfying results. The work of \cite{crafton2019direct} shows that sparse matrices of full rank allow training DFA on large datasets by reducing the memory cost of big matrices. Such matrices are a significant obstacle in DFA training on large datasets as observed in \cite{crafton2019direct, xiao2018biologically, moskovitz2018feedback, xiao2018biologically, moskovitz2018feedback}. The \cite{crafton2019direct} also transfer BP trained models to DFA training showing comparable results to BP. The sparse matrices can be used in parallel with the MEM-DFA described in this work. 

Biologically inspired algorithms are an active area of research, and many works try to improve the performance of biologically motivated methods. Our work does not focus on increasing the accuracy of FA or DFA but proposes an algorithm based on those methods, which reduces memory usage significantly during deep neural networks training.

\section{Backpropagation}

\begin{figure}[h]
  \begin{center}
  \includegraphics[width=0.8\columnwidth]{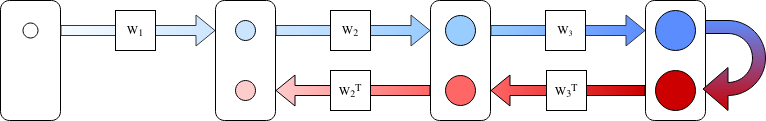}
  \end{center}
  \caption{Backpropagation. The gradient is propagated by transposed weight matrices.}
  \label{fig:bp}
\end{figure}

This section defines the notation and terminology, starting from the backpropagation algorithm for a sequential network with $n$ layers presented in Figure~\ref{fig:bp}. By layer, we usually mean an affine transformation together with one or more subsequent nonlinear operations. The operations within a layer are often called sublayers. 

We denote by $x$ or $a_0$ an input vector, $y$ an expected output, $f$ an activation function, $C$ a cost function, $W_i$ a weight matrix and $b_i$ a bias where bottom index $i \in \{1..n\}$ means the $i$-th layer. Forward calculations can be described by
\begin{align}
  z_i &= W_i a_{i-1} + b_i \\
  a_i &= f(z_i)
\end{align}
where $z_i$ is a weighted vector and $a_i$ is an activation vector. We call the vector $a_n$ the model's prediction. In the $i$-the layer, we consider gradients with respect to weights $W_i$, bias $b_i$ and vector $z_{i-1}$ and we denote them by $\pder{C}{W_i}$, $\pder{C}{b_i}$ and $\pder{C}{z_{i-1}}$, respectively.

For simplicity, the following cost function is considered
\begin{equation}
    C = \frac{1}{2}||y - a_n||^2\,.
\end{equation}
Hence, gradient with respect to $a_n$, also called error, is 
\begin{equation}
    \pder{C}{a_n} = a_n - y\,.
\end{equation}

By using the chain rule and by working backward it is possible to effectively calculate gradients with respect to weights and bias \cite{backpropagation}. The process can be described inductively. Firstly, based on the gradient with respect to the weighted vector $z_{i+1}$ we calculate the gradient with respect to $a_i$ through weight matrix $W_{i+1}$
\begin{equation}
  \label{eq:der-ai}
  \pder{C}{a_i} = W^T_{i + 1} \pder{C}{z_{i+1}}\,.
\end{equation}
Having $\pder{C}{a_i}$, we propagate through activation function and get gradient with respect to $z_i$ in result
\begin{equation}
  \label{eq:der-zi}
  \pder{C}{z_i} = \pder{C}{a_i} \odot f'(z_i)\,.
\end{equation}
Thus we calculate the gradients with respect to weights and bias
\begin{align}
  \pder{C}{W_i} &= \pder{C}{z_i} a^T_{i-1}\,, \label{eq:der-wi}\\
  \pder{C}{b_i} &= \pder{C}{z_i}\,.
\end{align}
These gradients allow us to update weights and bias, for example, by using stochastic gradient descent method.

The main reason for memory consumption in the backpropagation algorithm is the need to store activation vectors. Hence the memory usage is growing with the number of layers. More specifically, we need $z_i$ and $a_{i-1}$ in order to calculate gradients in $i$-the layer as shown in equations (\ref{eq:der-zi}) and (\ref{eq:der-wi}). However, before calculating gradients in $i$-th layer, the gradients in $i + 1$-th layer have to be calculated first. Therefore, before we can forget $z_i$ and $a_{i-1}$ we need to use $z_{i+1}$ and $a_i$ in calculations of gradients in the $i+1$-th layer. Thus, in the stack's peak moment, all activation and weighted vectors are kept in memory to calculate gradients in each layer efficiently.

\section{Feedback Alignment}
\begin{figure}[h]
  \begin{center}
  \includegraphics[width=0.8\columnwidth]{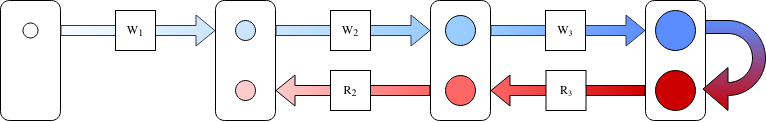}
  \end{center}
  \caption{Feedback Alignment. The error is propagated by random matrices instead transposed weight matrices}
  \label{fig:fa}
\end{figure}

The Feedback Alignment (FA) introduced in \cite{random, liao2016important} and presented in Figure~\ref{fig:fa} is a biologically-motivated algorithm. It targets the conjectured implausibility of the symmetric backward connectivity pattern in the backpropagation, i.e., the use of transpose of a weight matrix in the backward phase. The authors propose random matrices to propagate error instead and show that such an algorithm provides learning.

Since we propagate error using random matrices instead of transposed weight matrix, in FA we have vectors which are not strictly gradients but which play role of the BP's gradients $\pder{C}{v}$. We denote them by $\delta v$. The FA algorithm differs from BP by the equation (\ref{eq:der-ai}).
\begin{equation}
  \label{eq:der-ai-fa}
  \delta a_i = R_{i+1} \delta z_{i+1}
\end{equation}
where $R_i$ is random matrix with the same dimensions as $W^T_i$. The $R_i$ matrix can be generated once at the beginning or newly generated for each iteration. Moreover, by preserving the signs of matrix values in $R_i$, we can achieve better results as shown in \cite{liao2016important}.

Although there is information alleviation between the forward and backward pass, the memory usage is roughly the same as for backpropagation since we still need to store all intermediate vectors.

\section{Direct Feedback Alignment}
\begin{figure}[h]
  \begin{center}
  \includegraphics[width=0.8\columnwidth]{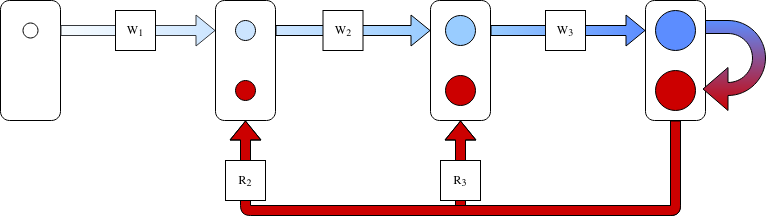}
  \end{center}
  \caption{Direct Feedback Alignment. The error is propagated directly from the last year by random matrices.}
  \label{fig:dfa}
\end{figure}

Direct feedback alignment (DFA) described in \cite{random2, random3} further develops the FA algorithm's idea.  A random matrix in DFA skips the dependence on the subsequent layer - it directly propagates the error from the last layer, as illustrated in figure \ref{fig:dfa}. Hence, the $R_i$ matrix has the input dimension of the model's output and output dimension as $W^T_i$. The DFA differs from BP and FA respectively by the equations (\ref{eq:der-ai}) and (\ref{eq:der-ai-fa})
\begin{equation}
  \delta a_i = R_{i+1} \delta a_n
\end{equation}
DFA still needs to store all intermediate vectors at the peak. Therefore it has similar memory complexity as BP and FA. However, DFA allows parallelizing some operations in the backward pass.

\section{Our method MEM-DFA}
Let us observe that since an error in DFA is propagated directly through the random matrix from the last layer, the gradient estimations are independent of gradients from the layers after. In other words, we could do the backpropagation independently from each layer. So assuming we have input to the first operation in the layer $i$ and error propagated through random matrix $R_i$, we could proceed with backpropagation within that layer without waiting for the gradient from the layer after.

Our method makes use of these observations. During the forward pass, we do not store any intermediate vectors.
Instead of one backward pass, we alternate the second forward pass and backward pass in the following way. We do forward pass within the layer, storing the intermediate vectors. Then we propagate the error directly to the last operation in that layer and then proceed with backpropagation within that layer. We do it starting from the first layer to use the output of the layer $i$ as input to the layer $i+1$. 

\begin{figure}
\begin{center}
\begin{tabular}{p{0.8\columnwidth}}
    \includegraphics[width=0.8\columnwidth]{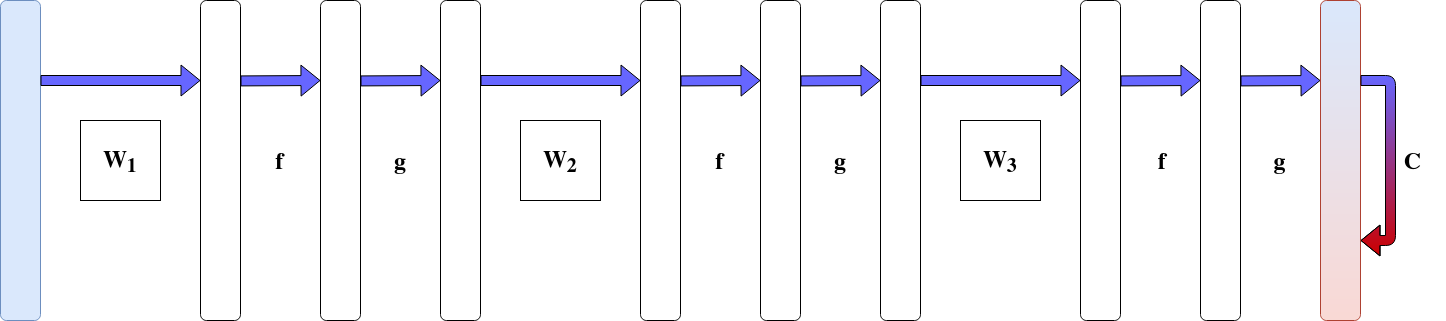}\label{fig:flow2a} \\
    \small{(a) During the first phase do not store vectors apart from input and output vectors.} \\
    \includegraphics[width=0.8\columnwidth]{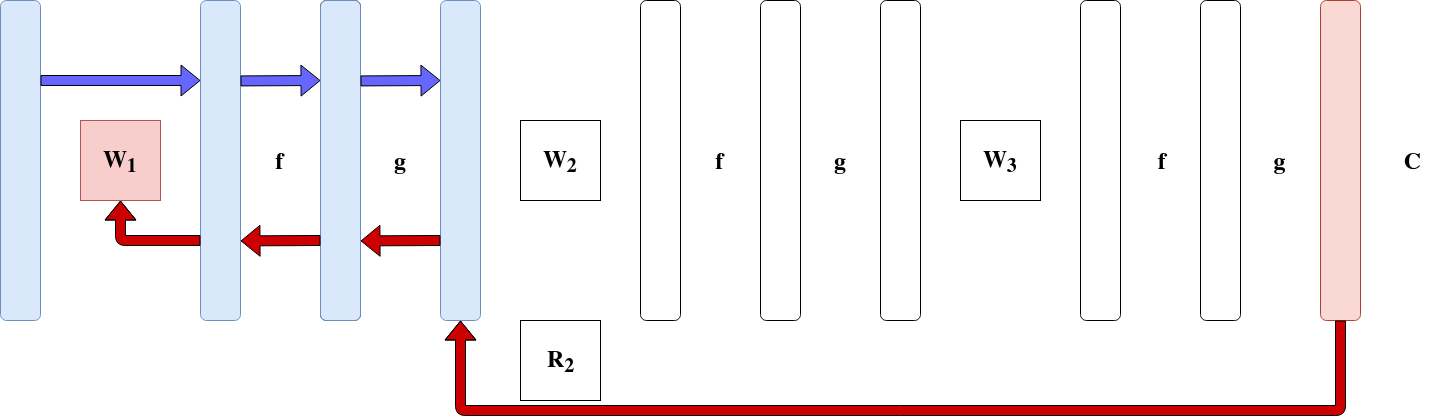}\label{fig:flow2b} \\
    \small{(b) Propagate error to the first layer through random matrix, calculate intermediate vectors within layer and store them. Then proceed with local backpropagation within layer.} \\
    \includegraphics[width=0.8\columnwidth]{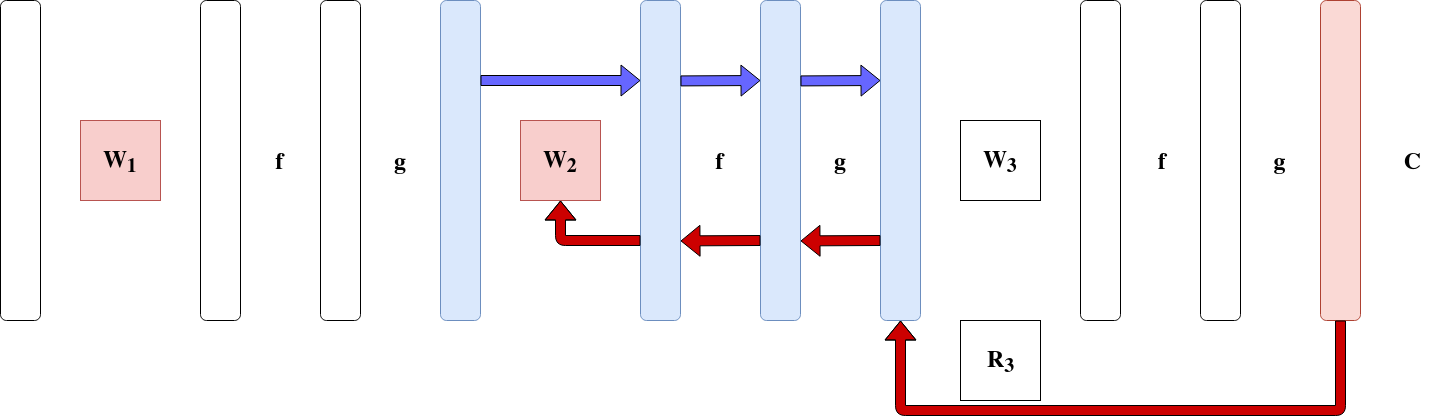}\label{fig:flow2c} \\
    \small{(c) Continue this process for the next layers by using the calculated activation vector from the previous layers.} \\
    \includegraphics[width=0.8\columnwidth]{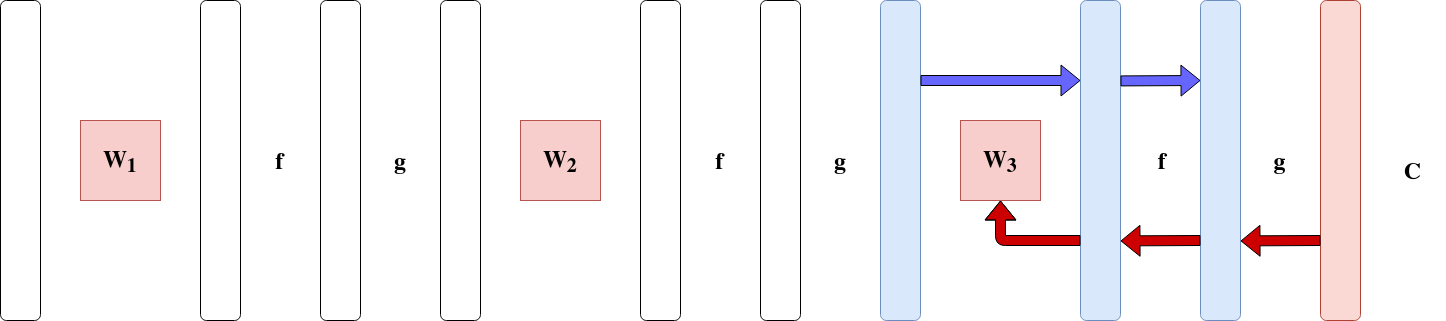}\label{fig:flow2d} \\
    \small{(d) For the last layer there is no need to propagate error through random matrix, just backpropagate within that layer.}
\end{tabular}
\end{center}
\caption{MEM-DFA algorithm. Each layer is composed of $W_i$, $f$, and $g$ operations. We can see that we do exactly two full forward and one full backward passes, but without the need to store all intermediate activations.} 

\label{fig:dfamem-flow}
\end{figure}

We called this method MEM-DFA. The algorithm is illustrated in figure \ref{fig:dfamem-flow}. We can describe it as follows.

\begin{algorithm}
\caption{MEM-DFA}
\begin{algorithmic}
    \STATE Calculate $a_n$ in the forward pass without storing intermediate vectors apart from input $a_0$.
    \STATE Calculate $\delta a_n$.
    \FOR{\texttt{$i \in \{1..n\}$}}
        \STATE 1. Using $a_{i-1}$ calculate $a_i$ in the forward pass within the $i$-th layer storing intermediate vectors.
        \STATE 2. Calculate $\delta a_i$.
        \STATE 3. Backpropagate locally within $i$-th layer and forget intermediate vectors apart from the $a_i$.
\ENDFOR
\end{algorithmic}
\end{algorithm}

Our method stores at most $k+1$ intermediate vectors, where $k$ is the number of operations in a layer. Usually, $k$ is insignificant compared to the number of layers. Thus, MEM-DFA achieves constant memory cost, regardless of the number of weight matrices in a neural network.

The computational cost of MEM-DFA is equivalent to two forward passes plus one backward pass of the DFA, FA, or BP. All operations are the same as in DFA but in a different order or eventually recalculated. The MEM-DFA method is numerically equivalent to DFA.

\section{Experiments}
\begin{table}
\begin{center}
    \begin{tabular}{||c|c|c|c|c|c||}
    \hline
    dataset & exp. no. & model & BP & FA & DFA / MEM-DFA \\
    \hline\hline
    MNIST & I & 3~x~FC & 97,6\% & 97,1\% & 97,1\% \\
    MNIST & II & 2~x~Conv + 2~x~FC & 99,0\% & 98,7\% & 98,9\% \\ 
    CIFAR-10 & I & 2~x~Conv + 2~x~FC & 70,1\% & 63,7\% & 64,3\% \\
    CIFAR-10 & II & 3~x~Conv + 2~x~FC & 74,8\% & 51,5\% & 51,2\% \\
    CIFAR-10 & III & VGG-16 & 77,8\% & 55,0\% & 54,8\% \\
    \hline
    \end{tabular}
\end{center}
\caption{The accuracy of the models trained with different learning methods on MNIST and CIFAR-10.}
\label{tab:accuracy}
\end{table}

Our experiments were conducted on MNIST and CIFAR-10 with various neural network models. Each model was trained with BP, FA, DFA, and MEM-DFA methods. We measured the memory usage and computation time per training iteration. The accuracy of our FA and DFA implementations were close to those from the original works \cite{random, liao2016important, random2}. The results are presented in Table~\ref{tab:accuracy}. Throughout the experiments, we used the ReLU activation function and softmax cross-entropy cost function. Weight updates followed the stochastic gradient descent algorithm. 

The first experiment has been conducted on MNIST with a model consisted of 3 fully connected layers of size 100, 30, and 10, respectively. We trained the model using a learning rate of 0.01, batch size of 100, and 100 epochs.

\begin{figure}[h!]
\begin{center}
\begin{tabular}{c c}
\includegraphics[width=0.35\columnwidth]{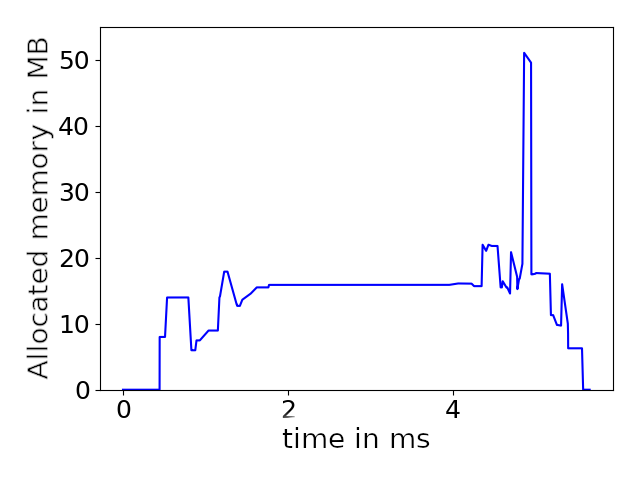} & \includegraphics[width=0.35\columnwidth]{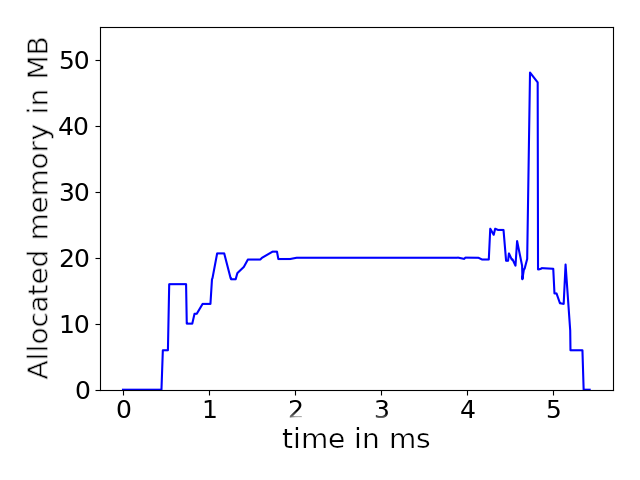} \\
\small{(a) Backpropagation} & \small{(b) Feedback Alignment} \\
\includegraphics[width=0.35\columnwidth]{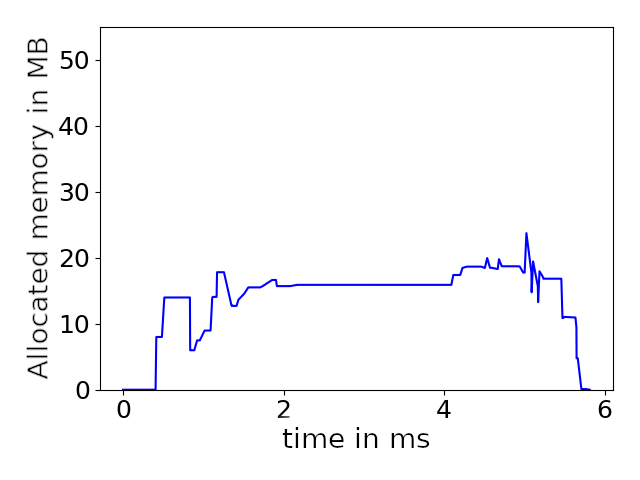} & \includegraphics[width=0.35\columnwidth]{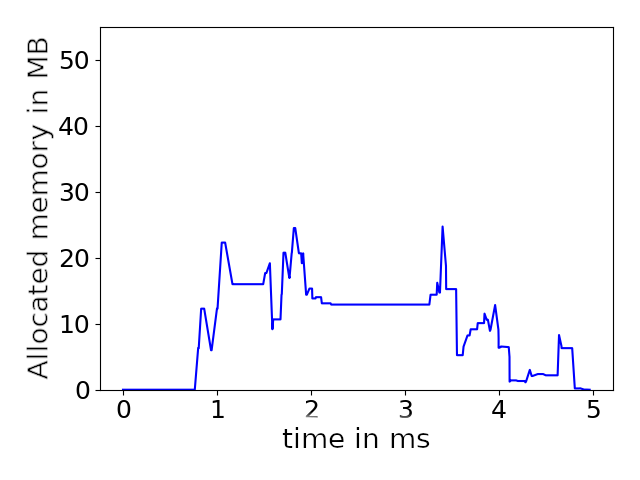} \\
\small{(c) Direct Feedback Alignment} & \small{(d) MEM-DFA}
\end{tabular}
\end{center}
\caption{Memory usage of convolution network on MNIST with 100 batch size.}
\label{fig:mnist-conv}
\end{figure}

In the next experiment on MNIST, we trained a convolutional neural network. The first convolution layer contained 20 filters of size $5x5$, and the second layer 50 filters of the same size. Additionally, the max-pooling of size $2x2$ and stride $2x2$ was applied after each Conv layer. In the end, there were two fully connected layers of sizes 500 and 10. The architecture is from \cite{liao2016important}. We trained for 150 epochs with a learning rate of 0.005. Measurements of memory usage of this model are compiled in figure \ref{fig:mnist-conv}.

\begin{figure}
\begin{center}
\begin{tabular}{c c}
\includegraphics[width=0.35\columnwidth]{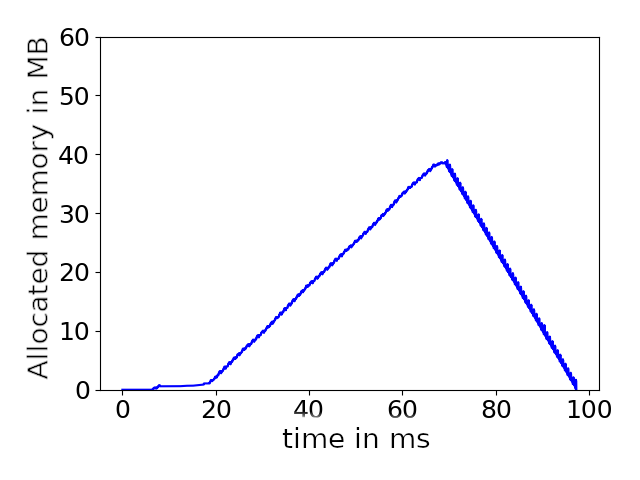} & \includegraphics[width=0.35\columnwidth]{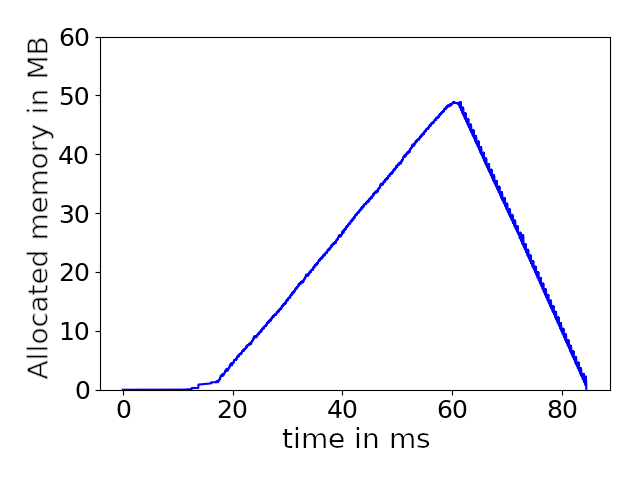} \\
\small{(a) Backpropagation} & \small{(b) Feedback Alignment} \\
\includegraphics[width=0.35\columnwidth]{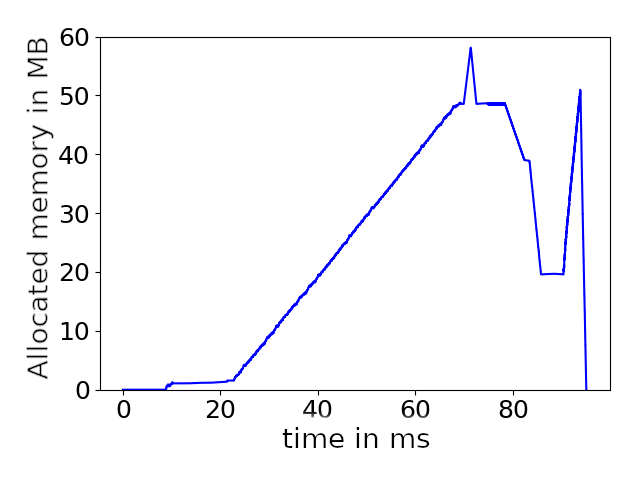} &
\includegraphics[width=0.35\columnwidth]{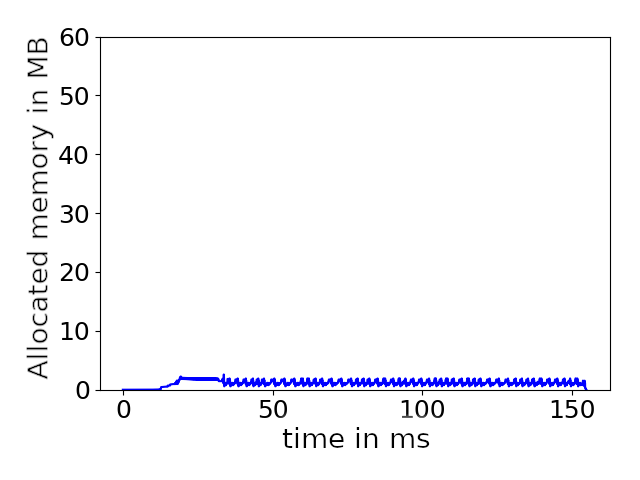} \\
\small{(c) Direct Feedback Alignment} & \small{(d) MEM-DFA} \\
\end{tabular}
\end{center}
\caption{Memory usage of the model with 50 FC layers on MNIST with 100 batch size.}
\label{fig:50layer}
\end{figure}

The final experiment on MNIST was conducted on a larger neural network with 50 fully connected layers, each of size 500. We trained with a batch of size 100. Figure \ref{fig:50layer} presents network's allocated memory with BP, FA, DFA, and MEM-DFA algorithms. This model was used only for memory measurements.

The next three experiments were conducted on CIFAR-10. In the first, we used the same CNN as before. Hyperparameters used were also very similar. 

In the second experiment, we used a model with convolutions with 32 filters of size $5x5$ in the first layer and 64 filters with the same dimensions in the next two. In the end, two FC layers with 128 and 10 sizes were used. The max and two average pooling of size $2x2$ and stride $2x2$ were put after respective Conv layers. The model also comes from \cite{liao2016important}.

\begin{figure}
\begin{center}
\begin{tabular}{c c}
\includegraphics[width=0.35\columnwidth]{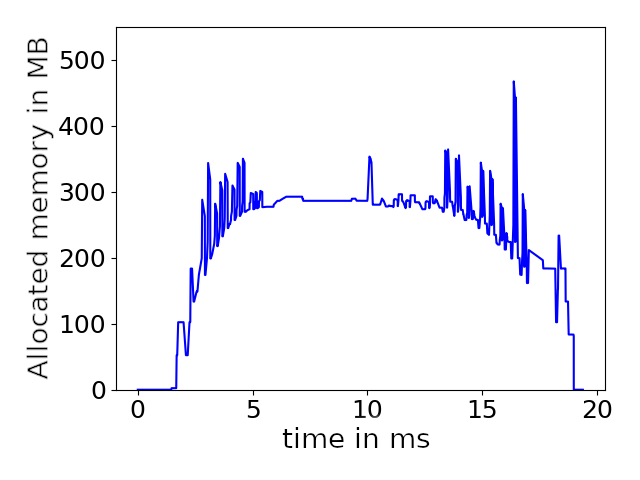} & \includegraphics[width=0.35\columnwidth]{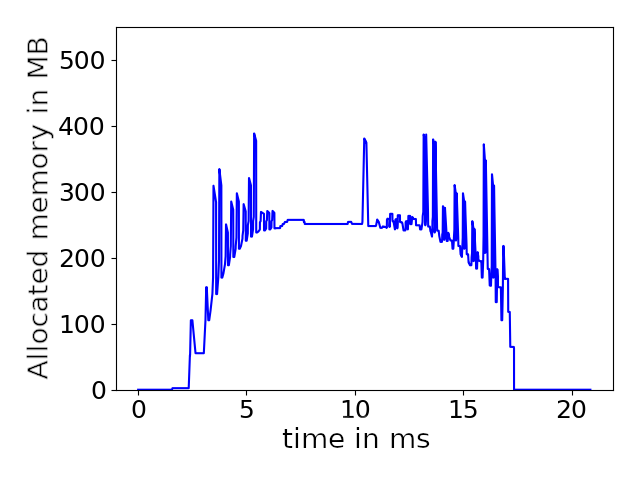} \\
\small{(a) Backpropagation} & \small{(b) Feedback Alignment} \\
\includegraphics[width=0.35\columnwidth]{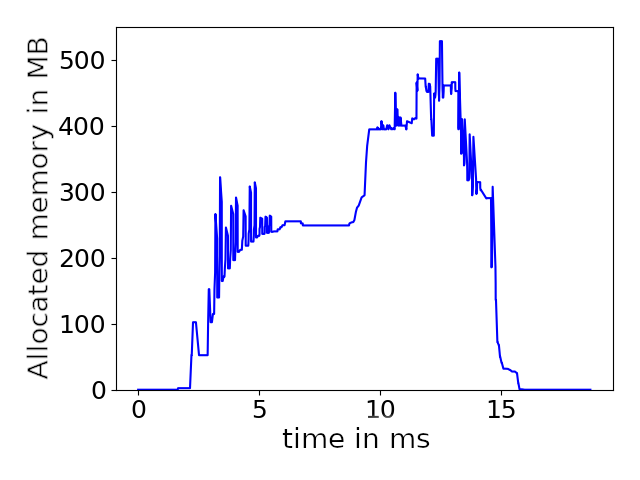} &
\includegraphics[width=0.35\columnwidth]{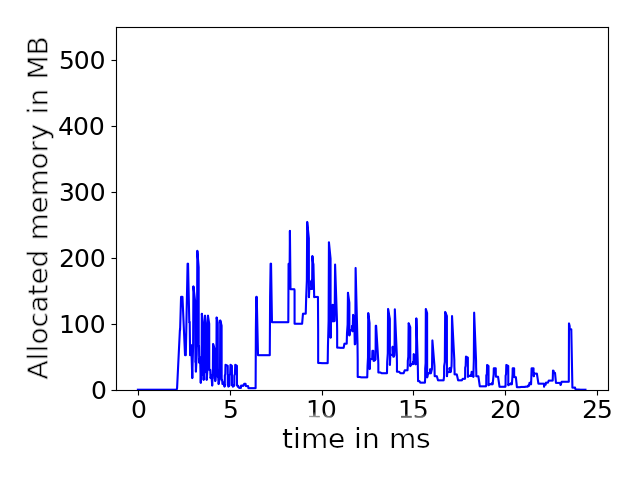} \\
\small{(c) Direct Feedback Alignment} & \small{(d) MEM-DFA}
\end{tabular}
\end{center}
\caption{Memory usage of VGG-16 on CIFAR-10 with 200 batch size.}
\label{fig:cifar-vgg}
\end{figure}

The VGG-16 network architecture \cite{simonyan2014very} was used in the final experiment on CIFAR-10. The models were trained using batch size 200 for 80 epochs with a learning rate of 0.001 for backpropagation and 0.00005 for FA and MEM-DFA. The memory measurements for various methods on VGA-16 are shown in figure \ref{fig:cifar-vgg}.

% \begin{table}[]
%     \begin{center}
%     \begin{tabular}{||c| p{0.7\columnwidth} ||}
%     \hline
%     name & architecture \\
%     \hline
%     \hline
%          fc  & FC(100), ReLu, FC(30), ReLu,  FC(10) \\
%          liao1 &   Conv((5,5), 20, (1, 1)), MaxPool((2, 2), (2, 2)), ReLu, Conv((5, 5), 50, (1, 1)), MaxPool((2, 2), (2, 2)), ReLu, FC(500), ReLu, FC(10) \\ 
%          liao2 &  Conv((5, 5), 32, (1, 1)), MaxPool((3, 3), (2, 2)), ReLu, Conv((5, 5), 64, (1, 1)), AvgPool((3, 3), (2, 2)), ReLu, Conv((5, 5), 64, (1, 1)), AvgPool((3, 3), (2, 2)), ReLu, FC(128), ReLu, FC(10)  \\ 
%          VGG-16 & - \\
%     \hline
%     \end{tabular}
%     \end{center}
%     \caption{Definitions of used architectures.}
%     \label{tab:my_label}
% \end{table}

\section{Measuring memory usage}
\subsection{Memory management in TensorFlow framework}

TensorFlow 1.12 \cite{tensorflow} has been used to create our neural network models. One of the main features of this technology is a static definition of a computation graph. The user defines the sequence of operations to be performed on input data at a later time. The computation graph is an abstract representation of the computation process, which allows preserving the relationship between input data and the results of applied operations. Thanks to it, it is possible to perform the symbolic computation of gradients. Such static construction allows the analysis, optimization, and compilation of the computation graph to utilize resources efficiently.

TensorFlow uses many techniques to minimize needed time to perform computation and used memory. XLA compiler \cite{xla} can replace consecutively appearing operations by optimized equivalent. It is also possible to detect repeating sequences of operations and introduce a common copy.

The memory management system of TensorFlow is implementing best-fit with the coalesce algorithm. It is reducing the fragmentation of memory and overall usage. TensorFlow chooses the smallest block during memory allocation, which still satisfies memory requirements from the memory block pool of predefined sizes. The block is kept alive until its reference counter has a positive value. When the counter becomes zero, all computations needed to access this particular memory are finished. When deallocating, the adjacent memory blocks are checked whether they are free, and if possible, they fuse into one chunk of a bigger size. Thus memory fragmentation is reduced.

Research on the memory management systems \cite{vramswap} has shown that calculating each memory fragment's liveliness based on the computation graph allows using a memory swapping strategy between GPU DRAM and RAM limiting peak usage of VRAM without sacrificing computation time.

Static definition of the computation graph allows for repeating calculations whose computational cost is smaller than keeping its results in memory for a prolonged time. Checkpointing algorithm introduced in \cite{reforwarding, sublinear} frees memory blocks if re-computation costs are below some specified threshold despite the positive reference counter for that memory block.

This research is limited to native TensorFlow optimizations. It allows measuring performance in a relatively easy to reconstruct manner. Many of the described above methods are perpendicular to each other and could be used simultaneously to limit memory usage significantly.

\subsection{Measuring memory usage}

The TensorFlow framework allows performing computations on CPU and GPU units. Unfortunately, its implementation cannot measure in detail allocated and deallocated RAM since messages created by allocator objects are stripped off requested memory size values. Tools such as htop, ps, valgrind, or docker containers with limited memory did not bring satisfactory results. TensorFlow is using a lazy strategy for deallocating memory to reduce interactions with the underlying operating system. When freeing a memory fragment, it is put back to memory poll without reporting to the operating system limiting these operations' overhead. Above all, CPU computations are usually used for fast prototype solutions. Thus this paper focused on measuring memory usage on GPU.

TensorFlow has a built-in operation to measure peak memory usage, which gives comparable results to other methods. Detailed time analysis is necessary to measure the proposed algorithm's performance, so this method was not used.

In the conducted experiments, Nvidia graphics cards have been used. Nvidia-Smi toolkit for this hardware allows observing card memory usage in real-time. Regrettably, TensorFlow is using a lazy strategy in memory management by maintaining an internal buffer of memory. It reduces interactions with firmware and operating system, which boosts the speed of running neural networks but blocks proper measurements from the outside tools like Nvidia-Smi, which shows constant memory usage during runtime and one deallocation at the end of it.

However, the GPU allocator in TensorFlow is emitting messages about the requested memory blocks. It is possible to extract information about total allocated memory in a given time and a chronological list of performed operations from the memory allocator's logs. To measure memory usage in our experiments, we adopted an existing code from \cite{checkpoinint_openai} on TensorFlow's memory optimization, which performs such data manipulations.

\section{Discussion}
Even on the shallow convolutional network on MNIST the memory decrease of MEM-DFA is apparent from figure \ref{fig:mnist-conv}. At the same time, the computational cost of one training iteration stays close to the other methods. The decrease in memory usage of MEM-DFA is further exemplified by VGG-16 on CIFAR-10, as shown by figure \ref{fig:cifar-vgg}. The computational time increased from about $20$ ms to about $25$ ms here. Moreover, we can also observe on this Figure the parallelization in the backward pass for DFA.

The memory usage and computational time of 50 layers fully connected model entirely agree with our theoretical expectations. In Figure $\ref{fig:50layer}$, we can see precisely the linear memory allocation during the forward pass and linear memory freeing during the backward pass of BP, FA, and partially DFA. In contrast, the memory usage of MEM-DFA stays roughly constant throughout the calculations. The computation time is about 50\% longer as predicted.

\section{Conclusion}
Biologically inspired algorithms in deep neural networks are an immensely active and fascinating area of research. Our work shows that they could provide insights regarding the nature of our brain and practical optimizations. 

The proposed MEM-DFA algorithm allows achieving constant memory complexity regardless of the number of linear operations in a neural network by using the higher independence between layers in DFA. The method increases the computational cost by a constant factor equal to one extra forward pass. Our experiments confirmed our theoretical results.

\bibliography{references}

\end{document}